\documentclass[runningheads]{llncs}
\usepackage{graphicx}
\begin{document}
\title{Well-Calibrated Probabilistic Predictive Maintenance using Venn-Abers\thanks{The authors acknowledge the Swedish Knowledge Foundation, Jönköping University, and the industrial partners for financially supporting the research through the AFAIR project with grant number 20200223, as part of the research and education environment SPARK at Jönköping University, Sweden.}}
\author{Ulf Johansson \and Tuwe Löfström \and Cecilia Sönströd}

\authorrunning{U. Johansson et al.}
\institute{Department of Computing, Jönköping University, Sweden \\
\email{\{ulf.johansson, tuwe.lofstrom, cecilia.sonstrod\}@ju.se}}

\maketitle              % typeset the header of the contribution
\begin{abstract}
When using machine learning for fault detection, a common problem is the fact that most data sets are very unbalanced, with the minority class (a fault) being the interesting one. In this paper, we investigate the usage of Venn-Abers predictors, looking specifically at the effect on the minority class predictions. A key property of Venn-Abers predictors is that they output well-calibrated probability intervals. In the experiments, we apply Venn-Abers calibration to decision trees, random forests and XGBoost models, showing how both overconfident and underconfident models are corrected. In addition, the benefit of using the valid probability intervals produced by Venn-Abers for decision support is demonstrated. When using techniques producing opaque underlying models, e.g., random forest and XGBoost, each prediction will consist of not only the label, but also a valid probability interval, where the width is an indication of the confidence in the estimate. Adding Venn-Abers on top of a decision tree allows inspection and analysis of the model, to understand both the underlying relationship, and finding out in which parts of feature space that the model is accurate and/or confident. 

\end{abstract}

\section{Introduction}
\label{Intro}
Manufacturers and other companies increasingly collect data from sensors in their factories and products. Predictive maintenance evaluates the condition of machinery and other equipment by performing offline or online condition monitoring. The goal of the approach is to decide when a maintenance activity is most cost-effective but before the equipment loses performance. The intended result is primarily a reduction in unplanned downtime costs because of failure, but also to lower the maintenance costs.
Recently, predictive maintenance based on data-driven methods applying machine learning, has become the most effective solution for smart manufacturing and industrial big data, especially for performing health perception, e.g., fault diagnosis and remaining life assessment \cite{zhang}. 

When performing fault detection, most data sets are very unbalanced, and it is the minority class (a fault) that is the interesting one. In fact, it could be argued that precision must be sacrificed for increased recall, i.e., more false alarms is acceptable, as long as most faults are in fact detected. With this in mind, many approaches directly target increasing the number of detected faults, using for instance cost-sensitive learning or different sampling techniques, see e.g., \cite{matzka2020explainable}.

In this paper, we investigate another approach, where we produce well-calibrated probabilistic predictors. The overall purpose is to provide the user with a correct estimation, on the instance level, of the probability that the instance represents a fault.

For the calibration, we suggest the use of \textit{Venn-Abers} predictors, which in addition to providing calibrated probabilities, are able to output well-calibrated probability intervals. In that setting, the width of an interval represents the confidence in the probability estimation. In the experiments, we first show the effects of Venn-Abers calibration, using different underlying models and comparing the results to existing alternatives. After that, we look specifically at the added informativeness when returning probability intervals instead of just a single probability estimate, differentiating between opaque and interpretable predictive models.

\section{Background}
\label{Background}
In this section, we first discuss probabilistic prediction in general, before briefly describing some well-known calibration techniques and providing an introduction to Venn prediction.
\subsection{Probabilistic Prediction}
A probabilistic predictor outputs not only the predicted class label, but also a probability distribution over the labels. Ideally, a probabilistic predictor should be \textit{valid}, i.e., the predicted probability distributions should perform well against statistical tests based on subsequent observation of the labels. 

In this paper, we focus on \textit{calibration}, i.e., we want:
\begin{equation} 
\label{e1.13} 
p(c_j \mid p^{c_j})=p^{c_j},
\end{equation} 
where $p^{c_j}$ is the probability estimate for class $j$. Specifically, in order to not be misleading, predicted probabilities must be matched by observed accuracy. As an example, if a number of predictions with the probability estimate $0.95$ are made, these predictions should be correct in about $95\%$ of the cases.  
While most predictive models are capable of producing probability estimates, there is no guarantee that these are well-calibrated. In fact, several models like naive Bayes \cite{niculescu-mizil-caruana2005} and decision trees \cite{ProvostDomingosPETS,Johsdm19} are notorious for being being poorly calibrated off-the-shelf. Recent studies show that even models assumed to be generally well-calibrated like modern (i.e., deep) neural networks \cite{pmlr-v70-guo17a} and traditional neural networks \cite{JohanssonIJCNN19} often are not. To rectify this, there exist a number of external methods for calibration, where Platt scaling \cite{platt1999} and isotonic regression \cite{zadrozny-elkan01} are the two most frequently used. Both these techniques, as well as Venn-Abers \cite{vovk2012vennabers}, normally utilize a separate labeled data set for the actual calibration.

\subsection {Platt Scaling and Isotonic Regression} 
\label{sec:PlattIso} 
Platt scaling \cite{platt1999} fits a sigmoid function to the scores obtained by the model. The function is
\begin{equation} 
\hat{p}(c \mid s)=\frac{1}{1+e^{As+B}},
\end{equation}
where $\hat{p}(c \mid s)$ is the probability that an instance belongs
to class $c$, given its score $s$. The parameters $A$ and $B$ are found by a
gradient descent search, minimizing the loss function suggested in \cite{platt1999}.

Isotonic regression \cite{zadrozny-elkan01}, is a non-parametric method, where an \textit{isotonic}, i.e., non-decreasing, step-wise regression function is used for generalized binning. The function is fitted to the model scores, minimizing a mean square error criterion, typically using the pair-adjacent violators algorithm. 

\subsection {Venn-Abers Predictors} 
\label{VennAbers}
Venn predictors \cite{vovk2004selfcalibrating} are multi-probabilistic predictors, i.e., they output $C$ probability distributions -- where $C$ is the number of classes -- with one of them, by construction, guaranteed to be valid, as long as the data is IID. The probabilities produced for a specific class label can be converted into a probability interval for that label, with the size of the interval indicating the confidence in the estimates, for details see e.g., \cite{johanssoncopa18}.

\textit{Venn-Abers predictors} are Venn predictors, i.e., they inherit the validity guarantees. Venn-Abers predictors are, however, in the basic form restricted to two-class problems since they assume that the underlying model is a \textit{scoring classifier}. After calibration, which is done by fitting two isotonic regression \cite{zadrozny-elkan01} models to the calibration set, each prediction is associated with a well-calibrated probability interval.

More formally, for each test object $x_{l+1}$ being predicted, an inductive Venn-Abers predictor is defined in the following way. 
\begin{enumerate}
\item Let $\{z_1, \dots, z_l\}$ be a training set where each instance ${z=(x,y)}$ consists of an \textit{object} $x$ and a \textit{label} $y$. 
\item Train a scoring classifier using a proper training set $\{z_1, \dots, z_q\}$ using $q<l$ instances. 
\item Produce prediction scores for the objects in the calibration set, including the test object, $\{s(x_{q+1}), \dots, s(x_l), s(x_{l+1})\}$. 
\item Fit two isotonic calibrators, $g_0$ and $g_1$, using \\ $\{(s(x_{q+1}),y_{q+1}), \dots, (s(x_l),y_l), (s(x_{l+1}),0)\}$,  and \\ $\{(s(x_{q+1}),y_{q+1}), \dots, (s(x_l),y_l), (s(x_{l+1}),1)\}$, respectively. 
\item Finally, let the probability interval for $y_{l+1}=1$ be $[g_0(s(x_{l+1})), g_1(s(x_{l+1}))]$.
\end{enumerate}

It should be noted that when an inductive Venn-Abers predictor is applied on top of a decision tree, all instances falling in the same leaf will obtain the same prediction score, so after calibration every leaf will contain a specific prediction, consisting of a label and an associated confidence (a probability interval). This very informative model can be inspected and analyzed like any decision tree.   

\section{Method}
\label{Method}
The data set used in the experiments was introduced in \cite{matzka2020explainable} with the purpose of providing the research community with a typical predictive maintenance data set. The data set is presented in detail in the original paper, but will be briefly described here as well. It consists of six input features representing quality (low/medium/high), air temperature in Kelvin, process temperature in Kelvin, rotational speed in rounds per minute (rpm), torque in Newton meter (Nm), and tool wear in minutes. The class labels are failure (minority class) and no failure. Out of 10 000 instances, 339 are failures, making the problem highly imbalanced. Failures could occur for various reasons, represented as three hidden binary features in the data set, i.e., these should not be used for the model learning. If one of these hidden binary features is active, then the instance is labeled as a fault.  We have transformed the quality feature into numeric format and of course removed the three hidden features describing the specific kinds of failures, but otherwise kept the data as is. 

In this paper, we have used \textit{decision trees} as the transparent models, since they provide acceptable predictive performance while still being interpretable. Decision trees are able to provide probability estimates for its predictions, given as the proportion of samples of the predicted class in a leaf. The decision tree implementation used is the \textit{DecisionTreeClassifier} in {\it scikit-learn}. 

As one of the opaque models, we used standard \textit{Random forests} \cite{breiman2001random}, i.e. ensembles of random trees, trained using bagging and restricted to a random subset of the available attributes in each split. When using the {\it RandomForestClassifier} in {\it scikit-learn}, the probability estimates of the forest is the mean predicted class probabilities of all trees, where the class probability of a single tree is the proportion of samples of the predicted class in a leaf. 

\textit{Extreme Gradient Boosting (XGBoost)} \cite{xgboost}, is a highly scalable implementation of \textit{gradient boosting} \cite{friedman2001greedy}. XGBoost includes several improvements over the original gradient boosting algorithm, such as weighted quantile sketch for approximate tree learning as well as enabling parallel and distributed computation of sparse data. The same calculation to get the probability estimate as the {\it RandomForestClassifier} is used by {\it XGBClassifier} in the {\it xgboost} package, implementing the algorithm in Python.

\textit{Logistic Regression} \cite{cox1958regression} is a regression method using the logit function to estimate the log-odds of a binary event based on a linear combination of one or more attributes. We include logistic regression in the experimentation since it, by construction, should produce well-calibrated probability estimates. The {\it scikit-learn} implementation of the algorithm was used with the \textit{max-iter} parameter set to 500. 

When obtaining a single probability estimate from the Venn-Abers probability interval $(p_0, p_1)$, for comparison with the other approaches, the suggestion in \cite{vovk2012vennabers} was used:
\begin{equation} 
\label{e1.17} 
p=\frac{p_1}{1-p_0+p_1}
\end{equation}
thus producing a regularized value with the estimate moved slightly towards the neutral value $0.5$. 

A number of metrics are used in the analysis. Predictive performance is analysed using accuracy, area under the ROC curve (AUC), precision and recall. 

The quality of the calibration was evaluated using the \textit{expected calibration error} (ECE). When calculating \textit{ECE}, the probability estimates for class 1 are divided into $M$, in this study ten, equally sized bins, before taking a weighted average of the absolute differences between the fraction of correct (\textit{foc}) predictions and the mean of the prediction probabilities (\textit{mop}). Here, $n$ is the size of the data set and $B_m$ represents bin $m$.

\begin{equation} 
\label{e1.68} 
\textrm{ECE} = 
    \sum_{m=1}^{M} 
    \frac{\left|B_m\right|}{n} 
    \Bigl| \textrm{foc}(B_m) - \textrm{mop}(B_m) \Bigr|
\end{equation}
For the evaluation, standard 10x10-fold cross-validation was used, so all results reported are averaged over the $100$ folds.%, making all counts $10$-times the original size of the dataset. 

\section{Results}
\label{Results}


Table \ref{tab:1} shows the predictive performance for decision trees, random forests and XGBoost, for Uncalibrated (\textit{Uncal}) models and after calibration using Venn-Abers (\textit{VA}), Platt scaling (\textit{Platt}) and isotonic regression (\textit{IsoReg}). Logistic Regression is also used as comparison. We see that both the random forest and XGBoost, as expected, are more accurate and have higher \textit{AUC} than decision trees and logistic regression. Overall, calibration has a minor impact on predictive performance, although the \textit{AUC} decreases slightly for all three calibrated models. The reason is the smaller training set when setting aside instances for calibration. 

Looking at precision and recall, the random forest has a precision of almost $0.90$ and a recall of approximately $0.62$, before calibration. Once calibrated, the precision goes down to between $0.83$ and $0.85$ and the recall up to between $0.63$ and $0.65$. This is expected, since random forests are generally underconfident in their probabilistic predictions. For the decision tree, however, the calibration has the opposite effect; the precision increases from $0.77$ to between $0.79$ and $0.81$, while the recall decreases from $0.65$ to between $0.57$ and $0.60$. This is consistent with the fact that decision trees tend to be overconfident in their probability estimates. For XGBoost, calibration reduces recall, similarly to the effect seen on decision trees, but a corresponding increase in precision is not evident. Logistic regression achieves very low performance, especially considering recall, and is only able to predict very few instances as positive. In the targeted scenario, this would not be acceptable, so we exclude logistic regression from further evaluation.  
\begin{table}[htbp]
\caption{Predictive performance}
\setlength{\tabcolsep}{6pt}

\centering
\footnotesize
\begin{tabular}{llccccccc}
\textbf{Model} & \textbf{Cal} & \textbf{Acc} & \textbf{AUC} & \textbf{Prec} & \textbf{Rec} & \textbf{\#PosPred} \\\hline
DT & Uncal & .981 & .904 & .768 & .645 & 2846 \\
 & VA & .981 & .902 & .790 & .595 & 2551 \\
 & Platt & .981 & .897 & .810 & .566 & 2369 \\
 & IsoReg & .981 & .897 & .796 & .588 & 2505 \\\hline
RF & Uncal & .985 & .966 & .895 & .616 & 2334 \\
 & VA & .984 & .962 & .831 & .648 & 2644 \\
 & Platt & .984 & .957 & .845 & .630 & 2529 \\
 & IsoReg & .983 & .959 & .834 & .634 & 2577 \\\hline
XGB & Uncal & .986 & .975 & .858 & .696 & 2752 \\
 & VA & .984 & .969 & .842 & .662 & 2664 \\
 & Platt & .984 & .932 & .861 & .647 & 2547 \\
 & IsoReg & .984 & .967 & .845 & .645 & 2590 \\ \hline
LR & Uncal & .970 & .894 & .705 & .196 & 941
\end{tabular}
\label{tab:1}
\end{table}

Turning to the calibration results, we present results for the overall calibration, i.e., the probability estimates for the minority class 1 (i.e., faults), but we look specifically at the predictions that are actually for the minority class. As explained above, the most important effect of the calibration would be to improve the probability estimates for the minority class predictions, i.e., having $p\geq0.5$ and consequently predicting a failure.

\begin{figure}[htbp]
\centering
		\includegraphics[trim={0.4cm 0.4cm 0.3cm 0.4cm},clip,width=0.65\linewidth]{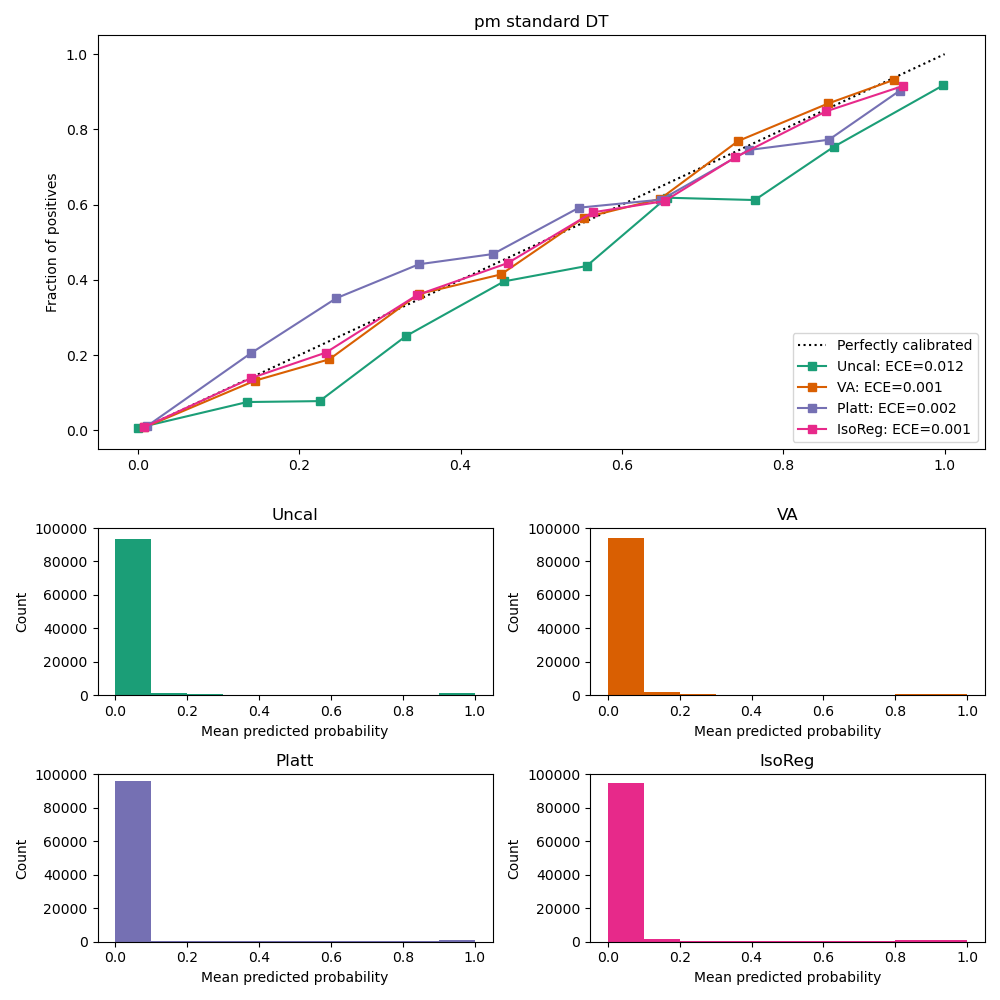}%
\caption{Reliability plot for all instances. Decision trees}
\label{fig:5}
\end{figure}

We now look at the reliability plots. Figure \ref{fig:5} shows an overall reliability plot when using decision trees. The decision tree appears to be excellently calibrated out-of-the box, with an \textit{ECE} of approximately $0.012$. However, the three calibration techniques all yield \textit{ECE:s} between $0.001$ and $0.002$. %In addition, the calibration significantly increases the \textit{ECE} to over $0.06$. But this is not the complete picture; what happens is that the calibration pushes the estimates towards the neutral value, which with such a dominant majority class will lead to a higher \textit{ECE}. 
Furthermore, when looking at all bins but the leftmost, we can see that the curves representing the calibrated values are much closer to the diagonal, especially for Venn-Abers and isotonic regression. We can also see that the trees are indeed overconfident in their probability estimates, as indicated by the green line being consistently below the dotted line.

\begin{figure}[htbp]
\centering
		\includegraphics[trim={0.4cm 0.4cm 0.3cm 0.4cm},clip,width=0.65\linewidth]{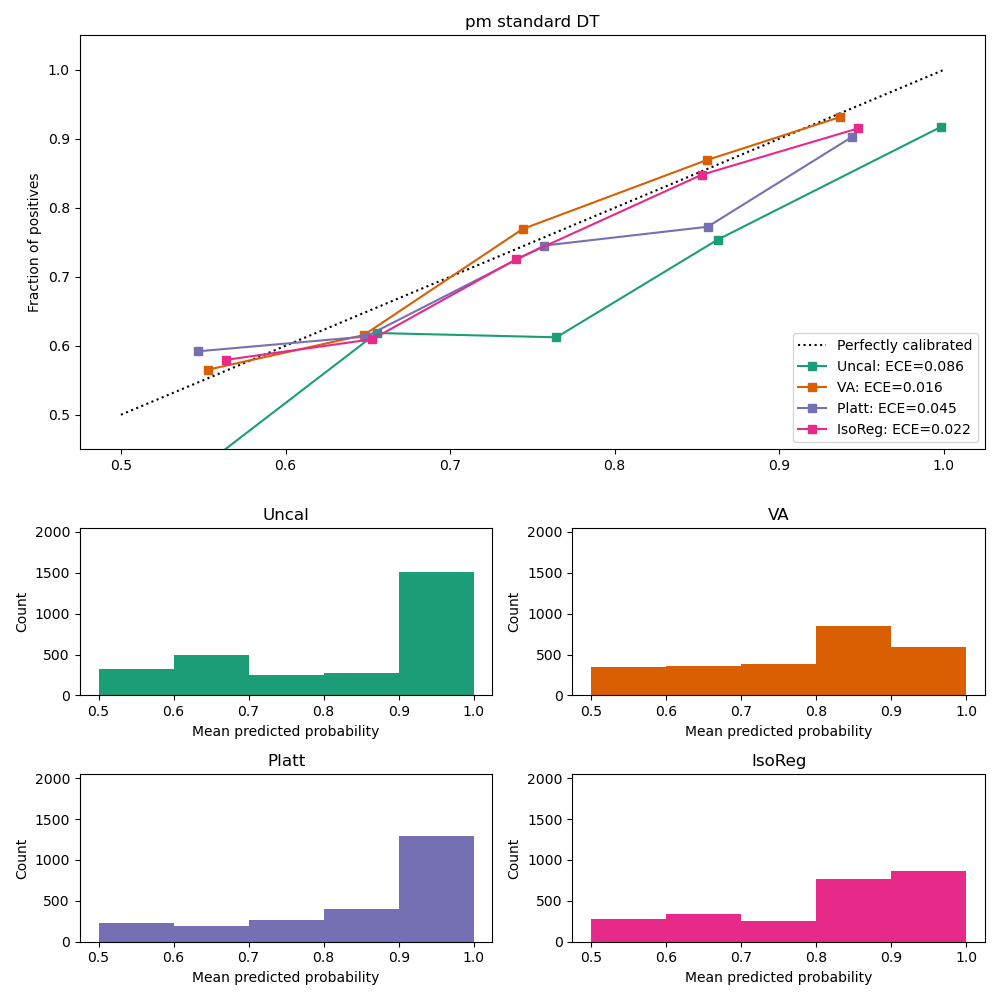}%
\caption{Reliability plot for minority class predictions. Decision trees}
\label{fig:6}
\end{figure}

Figure \ref{fig:6} shows the corresponding reliability plot when looking only at the minority class predictions. 
This is a very interesting plot, since it directly shows how calibration affects the minority class prediction, i.e., in the investigated scenario the predictions that would require an action. Here, it is clear that the decision tree is very overconfident in these predictions. In fact, after calibration, both Venn-Abers and isotonic regression result in very few predictions with a higher confidence than 0.9. Apart from Platt scaling, which is still slightly overconfident, calibrating the decision tree results in very good calibration without any clear tendency to over- or underconfidence. %Overall, the calibration makes the model slightly underconfident, instead of very overconfident, when it comes to predicting the interesting minority class. 
The \textit{ECE} drops from approximately $0.09$ to between $0.02$ and $0.05$.

\begin{figure}[htbp]
\centering
		\includegraphics[trim={0.4cm 0.4cm 0.3cm 0.4cm},clip,width=0.65\linewidth]{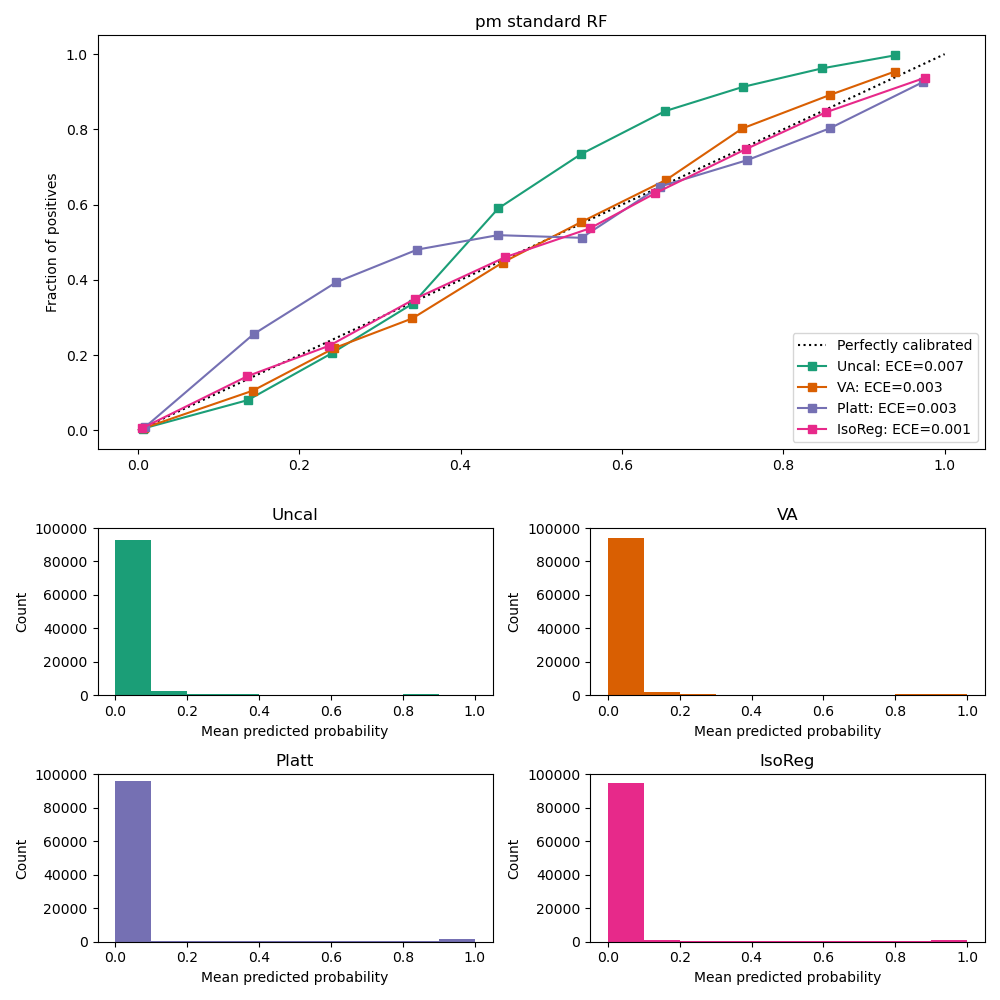}%
\caption{Reliability plot for all instances. Random forests}
\label{fig:7}
\end{figure}

\begin{figure}[htbp]
\centering
		\includegraphics[trim={0.4cm 0.4cm 0.3cm 0.4cm},clip,width=0.65\linewidth]{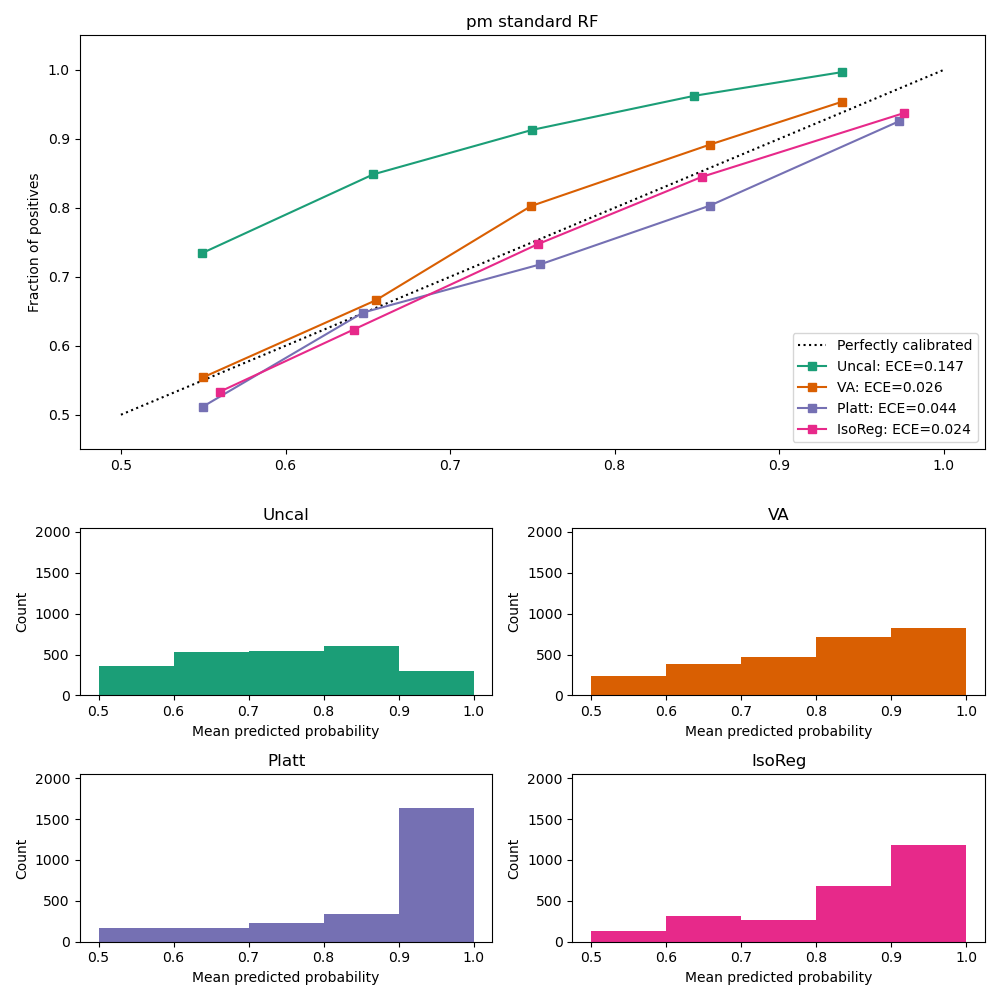}%
\caption{Reliability plot for minority class predictions. Random forests}
\label{fig:8}
\end{figure}

Turning to the random forests, Figure \ref{fig:7} shows the overall reliability plot.  Again, the probability estimates are well-calibrated, with an \textit{ECE} below $0.01$. Consequently, calibration has limited effect. 
However, when looking at the corresponding reliability plot for the minority class predictions only, in Figure \ref{fig:8}, it is seen that the random forest is in fact very underconfident when predicting the minority class. 
Again, calibration using Venn-Abers and isotonic regression are able to correct this, leading to an \textit{ECE} of $0.03$, compared to the original $0.15$. Platt scaling again improves the calibration compared to the underlying model, but to a lesser degree than the other calibrators.

\begin{figure}[htbp]
\centering
		\includegraphics[trim={0.4cm 0.4cm 0.3cm 0.4cm},clip,width=0.65\linewidth]{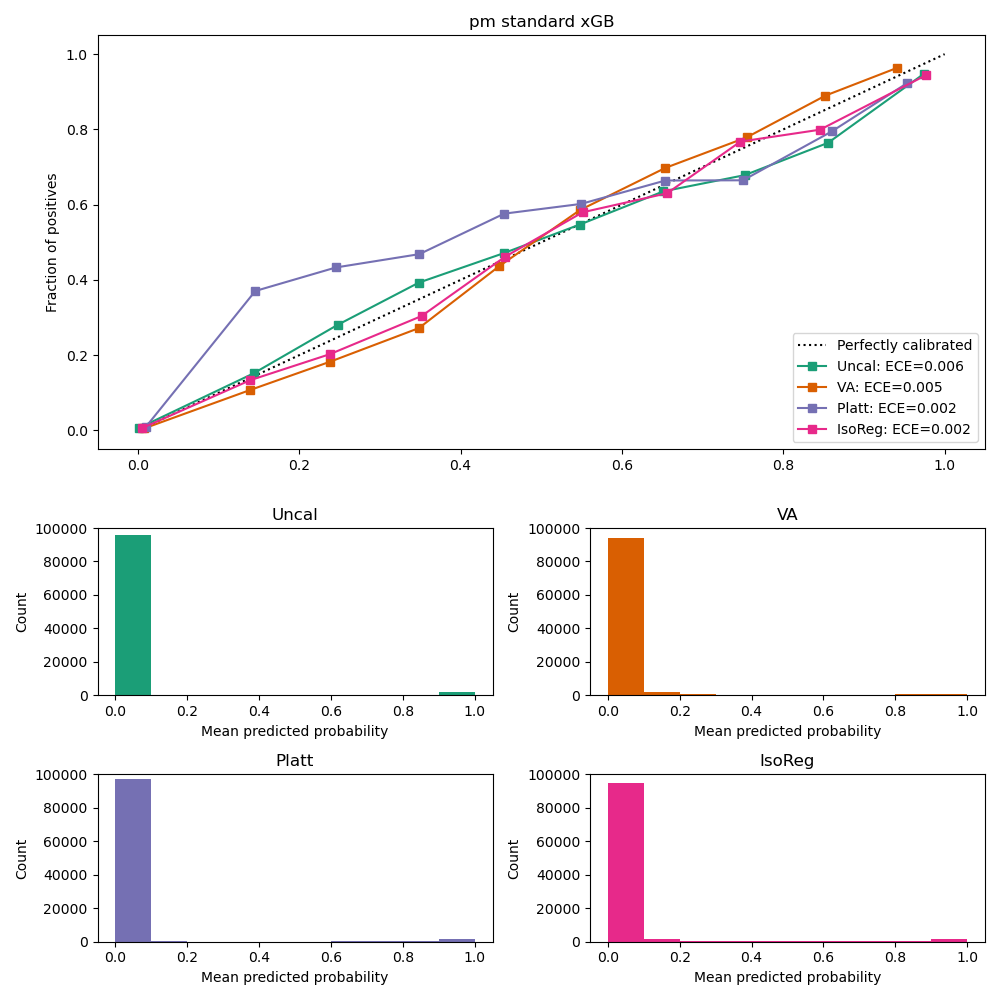}%
\caption{Reliability plot for all instances. XGBoost}
\label{fig:9}
\end{figure}

\begin{figure}[htbp]
\centering
		\includegraphics[trim={0.4cm 0.4cm 0.3cm 0.4cm},clip,width=0.65\linewidth]{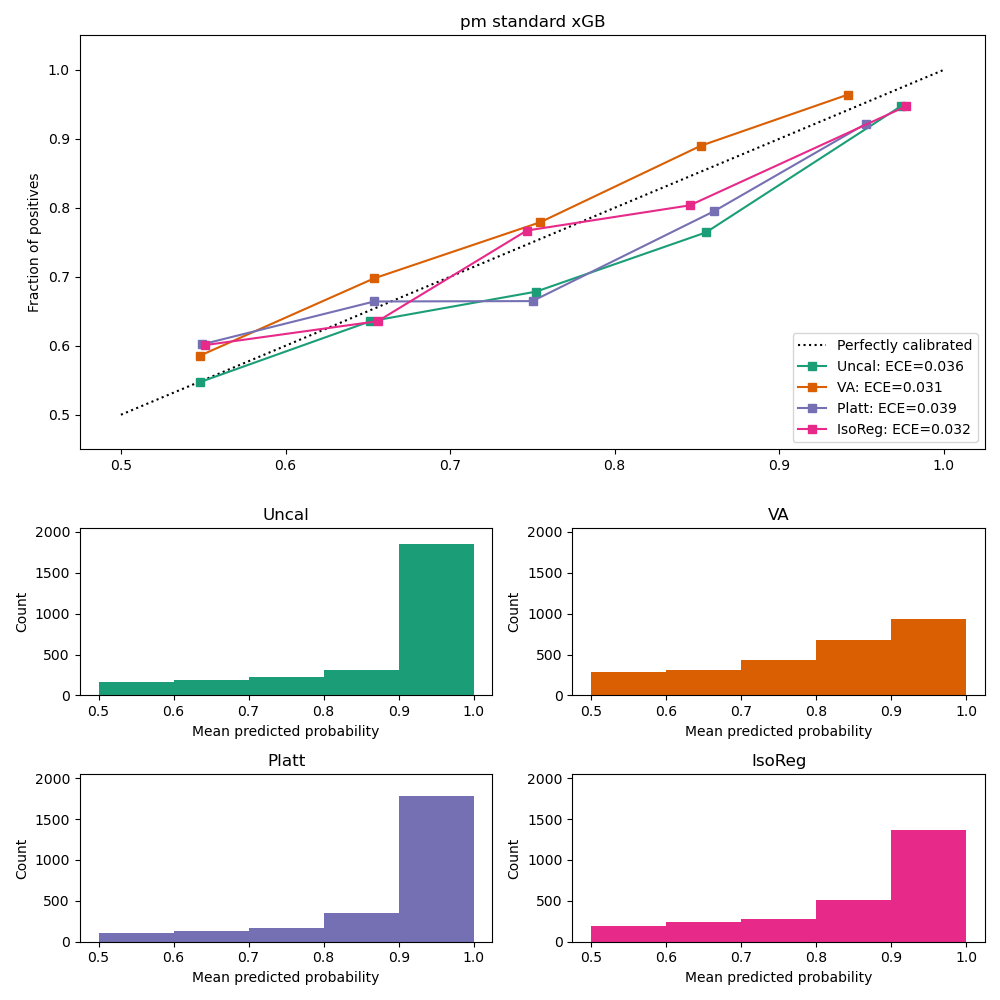}%
\caption{Reliability plot for minority class predictions. XGBoost}
\label{fig:10}
\end{figure}

Figure \ref{fig:9} shows the overall reliability for XGBoost. Again, the  calibration is excellent, with an \textit{ECE} below $0.01$, so further calibration has a very minor effect.  
When looking at the corresponding reliability plot for the minority class predictions only, in Figure \ref{fig:10}, it is seen that XGBoost is slightly overconfident when predicting the minority class. 
Even though XGBoost is fairly well calibrated on minority class predictions, both Venn-Abers and isotonic regression are able to slightly improve calibration. 

Table \ref{tab:2} summarizes the calibration results. ECE is the expected calibration error on all instances, and ECE-1 is the expected calibration error on the minority class predictions only. 
\begin{table}[htbp]
\caption{Expected Calibration Error - ECE}
\setlength{\tabcolsep}{6pt} 
\centering
\footnotesize
\begin{tabular}{lccc|ccc}
\textbf{} & \multicolumn{3}{c|}{\textbf{ECE}} & \multicolumn{3}{c}{\textbf{ECE-1}} \\
\textbf{Cal} & \multicolumn{1}{c}{\textbf{DT}} & \multicolumn{1}{c}{\textbf{RF}} & \multicolumn{1}{c|}{\textbf{XGB}} & \multicolumn{1}{c}{\textbf{DT}} & \multicolumn{1}{c}{\textbf{RF}} & \multicolumn{1}{c}{\textbf{XGB}} \\\hline
Uncal & .012 & .007 & .006 & .086 & .147 & .036 \\
VA & .001 & .003 & .005 & .016 & .026 & .031 \\
Platt & .002 & .003 & .002 & .045 & .044 & .039 \\
IsoReg & .001 & .001 & .002 & .022 & .024 & .032
\end{tabular}
\label{tab:2}
\end{table}

All-in-all, the experimental results show that applying calibration is very successful when looking at the minority class predictions. Interestingly, calibration is able to correct both the underconfident random forests and the overconfident techniques decision trees and XGBoost. Specifically, the probability estimates for the minority class predictions are much closer to the actual accuracy, i.e., they would provide a better decision support. 

So far we have treated Venn-Abers as a calibration technique, only using the regularized midpoint of the probability interval. However, for decision-making, there are two main reasons for why it makes a lot of sense to also use the probability intervals: i) they provide valid probability estimates of each class label and, ii) the width of the interval conveys insight into the uncertainty of the probability estimates. 

When using opaque models, Venn-Abers will provide probability intervals per instance for the positive class, as can be seen in Table~\ref{tab:rf}. The columns are $y=$ true target, $\hat{y}=$ prediction from the underlying model, $p_1=$ the probability estimate for the Failure class of the underlying model, $\tilde{y}=$ the prediction of the Venn-Abers after calibration, and finally $p_1^L$ and $p_1^H$ represent the probability interval for the positive class. 
\begin{table}[htbp]
\caption{Venn-Abers predictions from a Random Forest}
\setlength{\tabcolsep}{8pt} 
\footnotesize
\centering
\begin{tabular}{c||c|c||c|cc||}
    $y$ & $\hat{y}$ & $p_1$ & $\tilde{y}$ & $p_1^L$ & $p_1^H$ \\
    \hline
    1 & 1 & .92 & 1 & .97 & 1.0 \\
    1 & 1 & .78 & 1 & .85 & 1.0 \\
    % 0 & 1 & .65 & 1 & .81 & .87 \\
    0 & 0 & .47 & 1 (?) & .50 & .70 \\
    0 & 0 & .46 & ? & .41 & .70 
\end{tabular}
\label{tab:rf}
\end{table}

The examples shown in Table~\ref{tab:rf} illustrate how the interval of a Venn-Abers predictor provides both a valid probability interval and a level of certainty. The third and fourth instances are interesting examples, where the probabilities are close to $0.5$, thus indicating uncertainty about the label. In addition, the intervals are wide, showing that the probability estimate in itself is uncertain.  

\begin{figure}[htbp]
\centering
\includegraphics[trim={5.2cm 2.8cm 4.1cm 4.0cm}, clip,     width=0.75\linewidth]{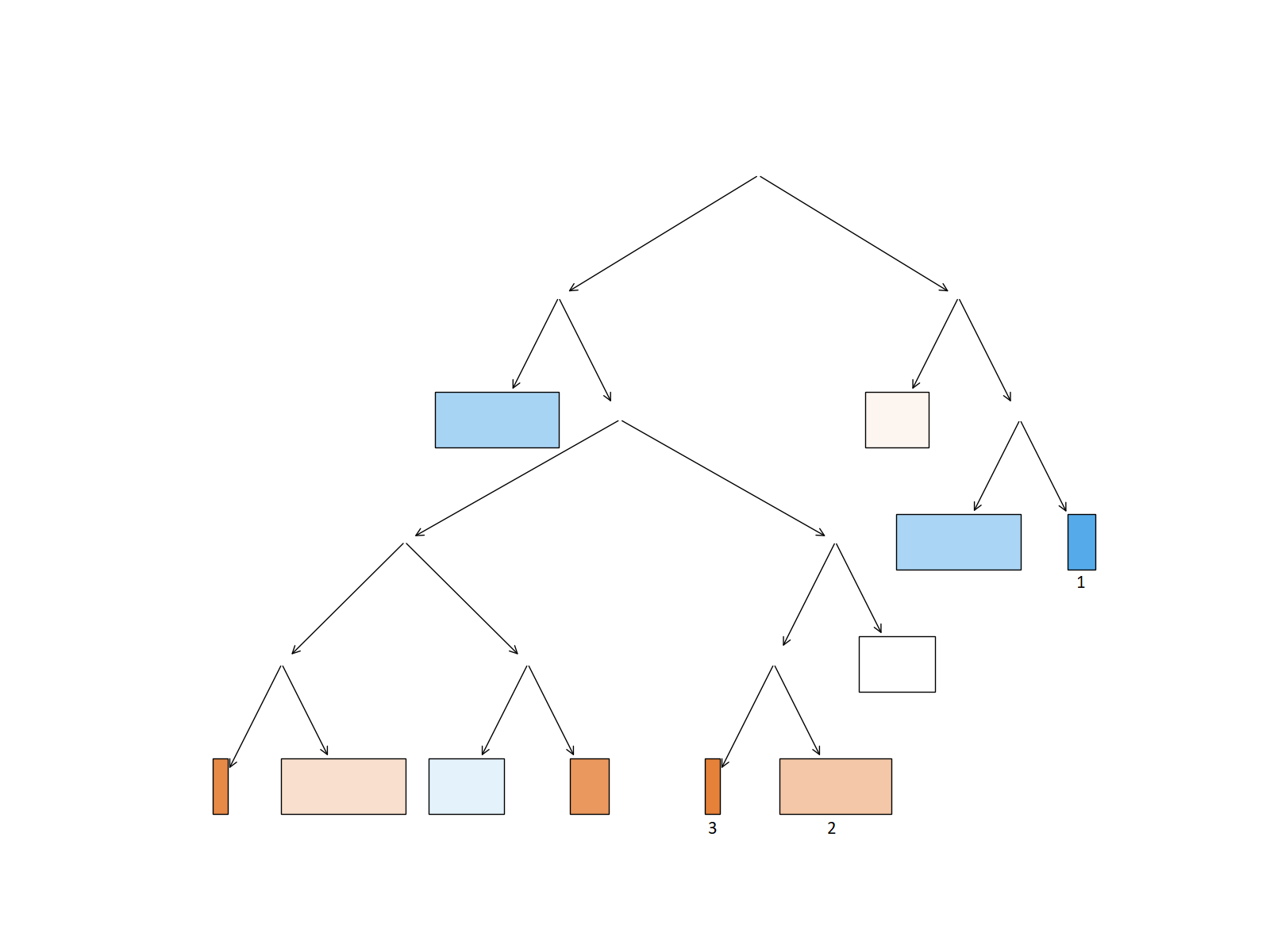}
    \caption{A well-calibrated and interpretable Venn tree. Colors represent the target classes with either no failure (orange) or failure (blue). The colour intensity corresponds to the estimated probability, while the width of the leaves gives the sizes of the Venn-Abers probability intervals, indicating how certain the probability estimates are.}
\label{fig:tree}
\end{figure}

When Venn-Abers is applied to a decision tree, the leaves will, after calibration, contain valid probability intervals, i.e., the tree can, in addition to providing explanations of the predictions, be inspected and analyzed to increase our understanding of the underlying relationship. Furthermore, by looking at the widths of the probability intervals in the leaves, we can identify different parts of feature space where the model is confident in its estimations and not. Fig.~\ref{fig:tree} shows a stylized Venn-Abers tree without the split criteria and the actual probability intervals. This tree is also pruned to a maximum depth of $5$ for visualization purposes. 

The Venn-Abers tree consequently conveys three things: 
\begin{itemize}
    \item The predicted class in the different leaves is given using different colors; blue and orange.
    \item The probability estimates are shown by colour intensity, with higher intensity representing higher probabilities. 
    \item The confidence in the estimates, represented by the width of the leaves.
\end{itemize} 

The right branch of the tree is more or less strongly dedicated to the prediction of failures, with some leaves being more certain about their probability estimates (smaller intervals and stronger color intensity), whereas other leaves are less certain about the correct class (lower color intensity), or indicate the model being less confident about its probability estimate (wider leaves). Most of the central and left parts of the tree are dedicated to the prediction of no failures with different degrees of probability and confidence, with some leaves having high probability with high confidence and others exhibiting a lot of uncertainty. The white and wide leaves indicate both a probability close to 0.5 and low confidence. 

Each leaf also represents a description, in the form of conjunctive conditions defined for the input attributes of the instances in that part of the feature space. Examples of rules describing the three numbered leaves in Fig.~\ref{fig:tree} are listed in Fig.~\ref{rules}. The interval at the end of each rule is the Venn-Abers interval for the probability.
\begin{figure}[htbp]
\footnotesize
\begin{eqnarray*}
   1) & & \textrm{torque [Nm] } > 67.4  \\ 
   & \& &  \textrm{rotational speed [rpm] } > 1239 \\ 
   & \rightarrow & \textbf{Failure } [0.91, 1.0]\\
   %2) & x_{1}>3.6 & \And \\ & x_{8}\leq0.34 & \And \\ & x_{4}\leq6.5 & \And \\ & x_{9}\leq3.5 & \And \\ & x_{5}>45.4 & \rightarrow \text{Blue } [0.993, 0.995]\\
 \cline{3-3}
%\\
   2) & & \textrm{torque [Nm] } \leq 59.05 \\ 
   & \& & \textrm{torque [Nm] } > 13.05 \\
   & \& & \textrm{rotational speed [rpm]  } >1380.5 \\ 
   & \& & \textrm{tool wear [min] } > 201.5 \\ 
   & \rightarrow & \textbf{No Failure } [0.62, 0.97]\\ 
 \cline{3-3}
%\\
   3) & & \textrm{torque [Nm] } \leq 59.05 \\ 
   & \& & \textrm{torque [Nm] } > 13.05 \\
   & \& & \textrm{rotational speed [rpm]  } >1380.5 \\ 
   & \& & \textrm{tool wear [min] } \leq 201.5 \\ 
   & \rightarrow & \textbf{No Failure } [0.98, 1.0]\\ 
\end{eqnarray*}
\normalsize
\caption{Rules for the three numbered leaves in Fig.~\ref{fig:tree}} 
\label{rules}
\end{figure}
Here, the first rule is very interesting, since it with high probability and confidence identifies faults. In a similar way, Rule 3, where a vast majority of all instances end up, identifies no faults. Rule 2, finally, points out a part of feature space where both the probability estimate and the confidence are rather low. Here, it could be noted that the only difference between this rule and Rule 3 is whether the value of the feature tool wear is higher or lower than $201.5$, suggesting that this could be looked into.     

\section{Concluding Remarks}
\label{CR}
We have in this paper demonstrated how Venn-Abers calibration could be used in a fault detection scenario with highly imbalanced classes. The experimental results show that calibration using either Venn-Abers or the common alternatives Platt scaling and isotonic regression, will lead to significantly improved minority class predictions. Specifically, predictions from both overconfident and underconfident models are corrected.    

In the second part, we argued for using Venn-Abers not only for calibration, but actually utilizing the added information present in the valid probability intervals produced. When used on top of an opaque model, each predicted label is complemented with a valid probability interval where the width indicates the confidence in the estimate. Adding Venn-Abers to a decision tree makes it straightforward to inspect and analyze the model in order to understand not only the underlying relationship, but also in which parts of feature space the model is strong and/or confident.   
%

\bibliographystyle{splncs04}
\bibliography{refs}

\end{document}